\documentclass{article}

\PassOptionsToPackage{numbers,compress}{natbib}

\usepackage[preprint]{neurips_2026}

\usepackage[utf8]{inputenc}
\usepackage[T1]{fontenc}
\usepackage{microtype}
\usepackage{amsmath,amssymb,amsfonts}
\usepackage{graphicx}
\usepackage{subcaption}

\usepackage{xspace}
\usepackage{dsfont}
\usepackage{algorithm}
\usepackage{algorithmic}

\usepackage{hyperref}       
\usepackage{url}            
\usepackage{booktabs}       
\usepackage{amsfonts}       
\usepackage{nicefrac}       
\usepackage{microtype}      
\usepackage{xcolor}         

\title{
Unsupervised Domain Shift Detection with Interpretable Subspace Attribution

}
\author{Anonymous Author(s)}

\author{
   Sebastian Springer \\
   SISSA \\
   \texttt{sspringe@sissa.it} \\
   \And
   Alessandro Laio \\
   SISSA \\
   \texttt{laio@sissa.it}
 }

\begin{document}

\maketitle

\begin{abstract}


We developed a tool for detecting domain shifts, namely subtle differences in the probability distributions of datasets. We identify these shifts using an algorithm designed to detect localised density anomalies in high-dimensional feature spaces. If an anomaly is present, we then identify the feature subspace in which the anomaly is most pronounced. This allows us to trace the domain shift to a small set of features, making the shift interpretable. Moreover, we provide a protocol for compensating domain shifts by extracting, from two unlabelled datasets, subsets of samples with no detectable residual distributional difference.
We validate the framework on controlled 20-dimensional benchmarks with known ground truth, recovering both broad and localized shifts together with their supporting feature subspaces. We then apply it to healthy electrocardiogram (ECG) recordings represented by 782 features. In age- and sex-matched cohort comparisons differing in measurement-device composition, the method detects device-induced shifts, extracts representative subsets enriched in the imbalanced device components, and identifies  ECG features associated with the acquisition contrast. These results suggest that density-shift detection and subspace attribution provide a practical framework for uncovering hidden cohort biases before downstream modelling.
\end{abstract}


\section{Introduction}

Comparing two collections of unlabeled samples is a fundamental task in machine learning. It arises in domain adaptation~\citep{quinonero2009dataset, sugiyama2012ml}, dataset auditing~\citep{rabanser2019failing}, covariate-shift correction~\citep{gretton2009covariate, sugiyama2008direct}, and the validation of learned representations~\citep{lopez2017revisiting}. The central statistical question is whether two finite samples are consistent with a common distribution and, when they are not, how the discrepancy is organized across samples and features.
 In machine-learning pipelines, undetected distributional discrepancies between training and deployment data, or between two reference cohorts, can distort learned representations, invalidate benchmark comparisons, and introduce systematic biases into downstream predictions~\citep{quinonero2009dataset,rabanser2019failing}. Identifying such discrepancies before fitting or evaluating downstream models is therefore an important component of reliable data-driven analysis. 

Global two-sample tests, such as maximum mean discrepancy~\citep{gretton2012kernel} and energy-distance tests~\citep{szekely2013energy}, provide evidence of distributional change at the dataset level. Classifier two-sample tests (C2ST)~\citep{lopez2017revisiting} cast the same problem as binary discrimination and can capture complex differences through flexible decision functions. These methods are powerful tools for testing or quantifying whether two datasets differ, but they do not, by design, return sets of samples associated with the excess mass in each dataset, nor do they  identify the feature subspaces that support the discrepancy.
In practice, distributional discrepancies  often involve a minority subpopulation, a restricted region of the sample space, or a feature subset in which the contrast is most pronounced, while the bulk of both datasets is comparable. In such settings, dataset-level tests may detect a discrepancy but provide limited information about its location or feature-level organization. Nearest-neighbor two-sample tests~\citep{schilling1986nn, friedman1979nn} exploit local geometry and can provide local evidence of distributional change, but they do not directly produce pruned/equalized subsets or a feature-weighted attribution of the detected discrepancy. Simultaneously localizing the responsible samples, attributing the discrepancy to an active feature support, and constructing equalized subsets therefore requires a more targeted pipeline.

We here introduce a framework for \emph{unsupervised detection, localization, and feature attribution} of distributional discrepancies between two unlabeled datasets. The framework  does not train a discriminative model to separate the two datasets. It proceeds in three stages. First, we use a recently introduced local two-sample score~\citep{EagleEye26} as a pointwise discrepancy backbone, assigning to each sample an anomaly score for local over- or under-representation relative to the other dataset. Second, we introduce a bidirectional tail-KS equalization procedure that identifies statistically significant excess-mass samples in each direction and constructs pruned subsets with undetectable discrepancy. Third, we introduce Neighbor-Enriched Subspace Localization, a procedure that identifies the active feature support by maximizing the anomaly score. 

We validate the framework on controlled 20-dimensional benchmarks with known ground truth. The method reliably recovers global shifts affecting extended subpopulations and, crucially, \emph{localized shifts} affecting structured minority populations. In the localized setting, a standard classifier two-sample baseline  is not able to identify the shift.
Having established recovery under controlled ground truth, we then apply the framework to healthy electrocardiogram (ECG) recordings from PTB-XL+~\citep{PTBxl,PTBxlplus}, represented by 782 clinically interpretable features. ECG cohorts provide a clinically relevant stress test for distributional discrepancy analysis because acquisition devices and measurement pipelines can induce systematic variation between nominally comparable cohorts. In age- and sex-matched comparisons constructed to differ primarily in measurement-device composition, our framework detects device-associated distributional shifts, extracts representative subsets enriched in the device components driving the mismatch, and identifies feature supports across controlled comparisons. These results suggest acquisition-related cohort bias rather than clinical distinction, illustrating how the framework can be used to audit dataset composition before downstream modelling.

The main contributions of this paper are:
\begin{itemize}
    \item \textbf{A bidirectional iterative equalization algorithm} that prunes excess-mass points from the two datasets (Section~\ref{sec:IDE-ZooM}).

    \item A \textbf{Neighbor-Enriched Subspace Localization} procedure that attributes the pruned set to the  features which are primarily responsible for the domain shift (Section~\ref{sec:neighbor_enriched}).

    \item \textbf{A complete domain shift detection and attribution pipeline}, requiring no class labels or supervised training on the datasets under comparison (Section~\ref{sec:method}).

    \item \textbf{Controlled synthetic experiments} with known ground truth demonstrating recovery of global and localized shifts and their supporting feature coordinates (Section~\ref{sec:20Dbenchmark}).

    \item \textbf{A real-world ECG application}  demonstrating detection of device-induced acquisition bias in age- and sex-matched healthy cohorts, with coherent feature attribution across controlled device-composition contrasts (Section~\ref{sec:ecg_dataset}).
\end{itemize}

\section{Method}
\label{sec:method}
\subsection{Backbone: EagleEye local two-sample scoring}
\label{sec:eagleeye_backbone}

We adopt EagleEye~\cite{EagleEye26} as a deterministic, unsupervised, training-free {two-sample} procedure that localizes
where two datasets differ in {density}.
Given two samples $\mathcal{X}=\{\mathcal{X}_1,\dots,\mathcal{X}_{n_{\mathcal{X}}}\}$ and  $\mathcal{Y}=\{\mathcal{Y}_1,\dots,\mathcal{Y}_{n_{\mathcal{Y}}}\}$,
EagleEye aims to identify \emph{localized density anomalies}: regions of feature space where
$\mathcal{Y}$ is significantly {over-} or {under-represented} relative to $\mathcal{X}$. 
We present the procedure for identifying anomalous points in $\mathcal{Y}$, but the procedure is
 run  in both directions to capture localized excesses and
deficits in each sample.
Let $\mathcal{U}=\mathcal{X}\cup \mathcal{Y}$ be the pooled set. For each test point $\mathcal{Y}_i\in \mathcal{Y}$, EagleEye computes its
nearest neighbours in $\mathcal{U}$ up to a maximal neighbourhood rank $K_M$ and encodes neighbour
memberships as a binary sequence
\[
b_i = (b_i^1,\dots,b_i^{K_M}),\qquad
b_i^k =
\begin{cases}
0, & \text{if the $k$-th neighbour of $\mathcal{Y}_i$ lies in $\mathcal{X}$},\\
1, & \text{if the $k$-th neighbour of $\mathcal{Y}_i$ lies in $\mathcal{Y}$}.
\end{cases}
\]
Under the null hypothesis that the two samples are \emph{locally exchangeable}, neighbour labels follow the global
mixture proportions, so the baseline probability that a neighbour belongs to $\mathcal{Y}$ is
$\hat p = \frac{n_{\mathcal{Y}}}{n_{\mathcal{X}}+n_{\mathcal{Y}}}$.
For $K\in\{1,\dots,K_M\}$, define the cumulative number of test-sample neighbours among the first $K$ neighbours as
\[
B(i,K)=\sum_{k=1}^{K} b_i^k.
\]
In the regime $K \ll n_{\mathcal{X}}+n_{\mathcal{Y}}$, we approximate $B(i,K)$ under the null as
$\mathrm{Binomial}(K,\hat p)$ and define an evidence score
\[
\Upsilon(i,K) := -\log\!\left(\Pr\!\left[B(i,K)\ge B_{\text{obs}}(i,K)\right]\right).
\]
The pointwise anomaly score is then obtained by maximizing across spatial scales,
\begin{equation}
\Upsilon_i := \max_{1\le K\le K_M}\Upsilon(i,K),
\end{equation}
yielding sensitivity to discrepancies at unknown spatial scales.

We flag putative anomalous points in the test sample as
\[
\mathcal{Y}^{+}:=\{\,\mathcal{Y}_i\in \mathcal{Y} \mid \Upsilon_i \ge \Upsilon^{*}_{+}\,\},
\]
where the critical threshold $\Upsilon^{*}_{+}$ is calibrated so that the \emph{per-point}
null exceedance probability equals a user-chosen level $p_{\mathrm{ext}}$ (default $10^{-5}$).
Because $\Upsilon_i$ is a maximum over $K\le K_M$, its null distribution is calibrated by
Monte Carlo using Bernoulli sequences of length $K_M$.
The flagged set $\mathcal{Y}^{+}$ can be interpreted as an anomaly map comprising an anomalous core together with a surrounding ``halo'' of points whose neighborhoods intersect a true discrepancy region.

\subsection{Tail-Based Bidirectional Equalization}
\label{sec:IDE-ZooM}

The next step of the procedure aims at identifying  the statistically significant excess mass relative to the corresponding backbone nulls of the observed anomaly-score distributions, and to remove such excesses jointly from the two samples whenever needed. This generalizes the density equalization procedure introduced in Ref.~\cite{EagleEye26}  making the criterion more directly tied to the global tail behavior of the score distributions and better suited to identify domain shift,  in which over- and under-densities may occur in nearby regions of space.
 For a fixed neighborhood depth $K_M$ and class proportion $p=n_{\mathcal{Y}}/(n_{\mathcal{X}}+n_{\mathcal{Y}})$, the null distributions for over- and under-density scores are estimated by generating a large number of Bernoulli sequences of length $K_M$ with parameters $p$ and $1-p$, respectively, and computing the corresponding anomaly scores as in Section~\ref{sec:eagleeye_backbone}. For each direction, we then define a tail threshold $\tau_{\mathrm{tail}}$ as a high null quantile (by default, the $0.97$ quantile) and compare the observed and null score tails above $\tau_{\mathrm{tail}}$ by means of a one-sided two-sample Kolmogorov--Smirnov test. We use a one-sided test because the goal of equalization is to reach a regime in which the empirical score tail is stochastically dominated by the corresponding null tail, meaning that extreme anomaly scores are no more frequent than expected under the backbone null. At each iteration, pruning is therefore activated on a given side only if the corresponding tail test rejects this condition, i.e., only if the observed score tail still exhibits a statistically significant excess relative to the matching null tail.

When at least one of the two tail tests rejects the null hypothesis, we perform a bidirectional equalization step. On each active side, it select at both tails the most  anomalous points (or, for computational efficiency, a small set of points from the corresponding score tail) and add them to the pruned representative sets $\hat{\mathcal{Y}}^{+}$ and/or $\hat{\mathcal{X}}^{+}$. The associated excess-mass neighborhoods are then removed from the current equalized samples, stopping at the first point belonging to the opposite sample, as in the implementation in Ref.\cite{EagleEye26}. In contrast to the procedure in Ref. \cite{EagleEye26}, however, this pruning is assessed jointly in the two directions, so that the neighborhood structure is updated while accounting for the total mass already removed from both samples. This is especially useful when over- and under-density modes lie in nearby regions of space.

To keep the procedure computationally efficient, it alternates between  local pruning  and a global score recomputation. During the pruning phase, equalization is carried out only with respect to the tail candidates identified at its start. After each pruning step, anomaly scores are recomputed only for the points belonging to the two tails, since these are the only ones needed to determine the next candidates for pruning. Once both tails have been equalized, anomaly scores are recomputed on the full equalized samples $\mathcal{X}_{\mathrm{eq}}$ and $\mathcal{Y}_{\mathrm{eq}}$, and the whole procedure is repeated until both KS tests pass. This outer iteration is important because pruning alters the neighborhood structure, so points that were initially outside the tails may enter them after the equalization step, while recomputing all scores after every pruning step would be computationally inefficient.
Pseudocode for the equalization procedure is provided in the Supplementary Material, Sec.~\ref{sec:Supplement_IDE-ZooM}.

\subsection{Neighbor-Enriched Subspace Localization}
\label{sec:neighbor_enriched}

In the final stage, the points extracted as described above are used to identify a subset of coordinates which determines the domain shift (if this subset exists).

Let $\mathcal{U}\in\mathbb{R}^{n\times d}$ denote the pooled feature matrix, let $\mathcal{Q}\subset\mathcal{U}$ be the query set, consisting of the high-evidence candidates associated with a domain shift identified as described above. Let $\mathcal{T}\subset\mathcal{U}$ be a target set, chosen as either $\mathcal{Y}$ (if $\mathcal{Q}\subset\mathcal{Y}$) or $\mathcal{X}$ (otherwise). We seek a feature weighting under which the $K$ nearest neighbors of the queries are maximally enriched in samples from $\mathcal{T}$. We introduce one nonnegative weight per feature. We distinguish between \emph{raw} weights, which are the directly learned nonnegative parameters, and \emph{effective} weights, which define the distance metric. The effective weights are obtained by rescaling the raw weights to scale at a constant d at each iteration to fix their overall magnitude, thereby removing the otherwise irrelevant global scale of the metric and preventing drift toward uniformly large or small weights. 
Distances are computed as,
\[
D_{w_{\mathrm{eff}}}(u_i,u_j)=\|\,w_{\mathrm{eff}}\odot u_i-w_{\mathrm{eff}}\odot u_j\,\|_2^2,
\]
whereas the sparsity-inducing penalty is applied to the raw weights. This decouples scale control from sparsity promotion: normalization fixes the overall weight budget of the metric, while the penalty remains free to favor solutions concentrated on a relatively small number of features.
The weights are learned through a differentiable soft $K$NN construction on stratified mini-batches $\mathcal{B}\subset\mathcal{U}$, sampled to maintain a fixed proportion of query and non-query samples. For each batch, weighted distances are converted into similarity logits,
\[
\ell_{ij}=-\,\frac{D_{w_{\mathrm{eff}}}(u_i,u_j)}{\tau},
\]
and neighbor probabilities are obtained by applying a row-wise softmax restricted to the $K$ largest logits in each row:
\[
P_{ij}=\frac{\exp(\ell_{ij})}{\sum_{j'\in \mathrm{TopK}(i)}\exp(\ell_{ij'})}
\;\;\text{for}\;\; j\in\mathrm{TopK}(i),
\qquad
P_{ij}=0\;\;\text{otherwise}.
\]
For each query $i\in \mathcal{Q}\cap \mathcal{B}$, we define the soft target mass
$
m_i=\sum_j P_{ij}\,\mathds{1}\{j\in \mathcal{T}\cap \mathcal{B}\}.
$.
The weights are then learned by maximizing the average soft target mass over the batch queries together with an $\ell_1$ penalty on the raw weights:
\[
\max\ \frac{1}{|\mathcal{Q}\cap \mathcal{B}|}
\sum_{i\in \mathcal{Q}\cap \mathcal{B}} m_i
\;-\;
\lambda \|w_{\mathrm{raw}}\|_1.
\]


After optimization, features are ranked by their effective weights.
This yields a compact set of coordinates that most strongly supports the localized discrepancy identified by the procedure in Section \ref{sec:IDE-ZooM}. For further details see Supp.~\ref{sec:supp_c1}.

The overall domain-shift identification protocol proceeds by alternating anomaly localization in the current feature space with feature-subspace refinement around the detected modes. Once a discrepancy has been localized, the associated pruned points are partitioned by mode, and Neighbor-Enriched subspace localization is applied to each mode separately. This mode-wise refinement is crucial because distinct modes may be supported by distinct subsets of coordinates. By isolating them individually, the procedure can reveal lower-dimensional subspaces in which the corresponding shift is sharper, more interpretable, and therefore easier to equalize selectively. The refinement can be iterated until the selected subspaces stabilize or a user-defined maximum number of iterations is reached.

\section{Experiments}
\label{sec:Experiments}

The proposed workflow is broadly applicable across domains in which localized distributional discrepancies and systematic shifts may coexist. Here, we first validate it in a controlled synthetic benchmark and then focus on a medically relevant application based on ECG-derived features.

\subsection{Controlled 20D benchmark for global and local domain-shift recovery}
\label{sec:20Dbenchmark}

We evaluate the proposed workflow in a controlled 20-dimensional benchmark with known ground truth. The reference distribution is a four-component Gaussian mixture, and each cohort contains $50{,}000$ background samples. On top of this common background, we consider two complementary types of domain shift. In the first experiment, we introduce a \emph{global} shift by displacing one Gaussian-mixture component along coordinates $(0,1,3)$, with amplitude controlled by a parameter $\sigma$. In the second experiment, we introduce a \emph{localized} shift by injecting into one cohort a structured minority population supported on coordinates $(2,4,6,8,9)$. Full construction details are provided in Supp.~\ref{supp:20d_benchmark_construction}.
This benchmark is designed to test two distinct capabilities of the method: first, whether it can recover a broad domain shift affecting an extended portion of the data and correctly identify the coordinates supporting it; and second, whether it can recover a localized minority shift embedded in a much larger background population and again localize the relevant coordinates. In both experiments, we repeat the analysis across multiple random seeds to assess the stability of both recovery and subspace identification.

\paragraph{Global domain-shift recovery.}

\begin{figure*}[t]
    \centering
    \includegraphics[width=.9\textwidth]{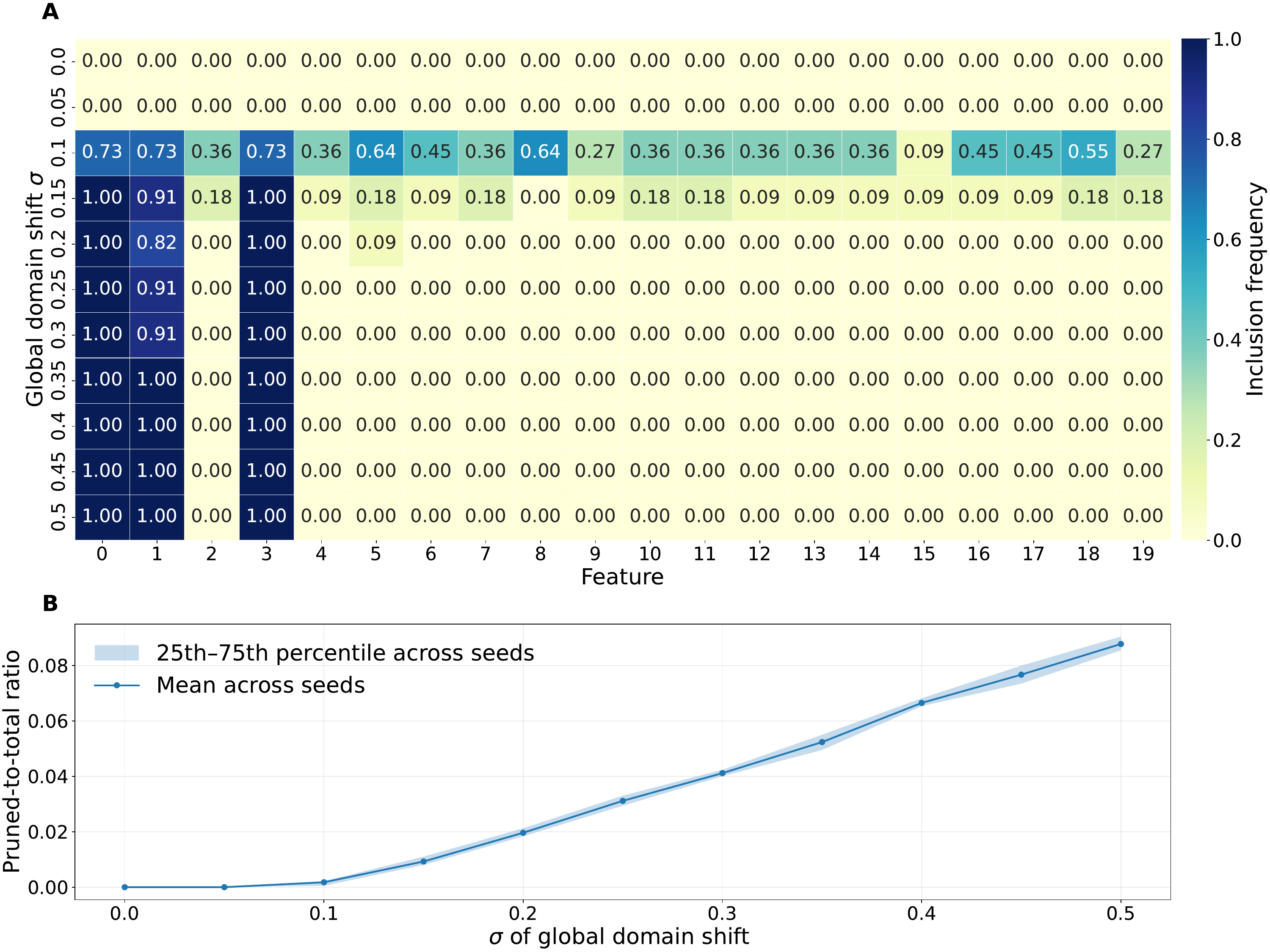}
\caption{
\textbf{Robustness of the identified shift set across global domain-shift amplitudes and random seeds.}
\textbf{(A)} Inclusion frequency of each feature in the identified shift set, aggregated across random seeds, as a function of the global domain-shift amplitude $\sigma$.
\textbf{(B)} Pruned-to-total ratio as a function of $\sigma$. The solid line denotes the mean across seeds, and the shaded region denotes the 25th--75th percentile range across seeds.
}
    \label{fig:global_shift_robustness1}
\end{figure*}

The result of the analysis for the global shift is illustrated in Fig.~\ref{fig:global_shift_robustness1}.  The shift amplitude is varied from $\sigma=0$ to $\sigma=0.5$ in increments of $0.05$, and each configuration is repeated across 11 random seeds.  For $\sigma=0$ and $\sigma=0.05$, the identified shift set is empty across seeds, indicating that the procedure does not produce spurious detections in the near-null regime. At $\sigma=0.1$, the true supporting coordinates $(0,1,3)$ already begin to emerge, although some non-supporting coordinates are still selected intermittently. By $\sigma=0.15$, the recovered set becomes visibly cleaner, and from $\sigma=0.2$ onward the method enters a stable regime: panel A shows that coordinates $(0,1,3)$ dominate the identified shift set with high inclusion frequency, while spurious coordinates are almost entirely eliminated. Consistently, panel B shows that the pruned-to-total ratio increases steadily with $\sigma$, indicating that broader global mismatches are increasingly easy to isolate through pruning. Taken together, these results show that the workflow can reliably detect a global domain shift and recover its true supporting coordinates once the shift exceeds a minimal detectable amplitude.

\paragraph{Localized domain-shift recovery}

\begin{figure*}[t]
    \centering
    \includegraphics[width=.9\textwidth]{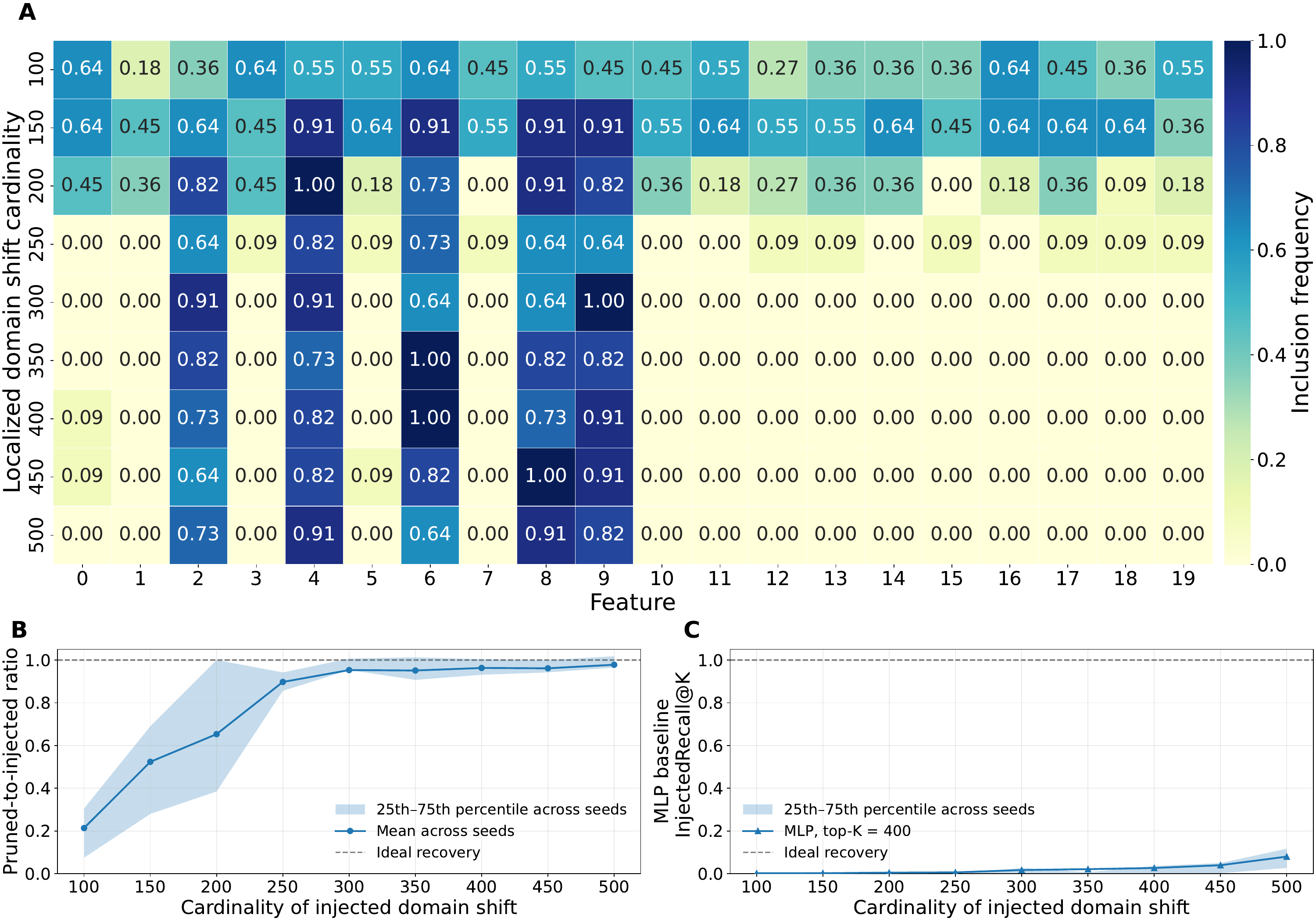}
\caption{
\textbf{Robustness of localized domain-shift recovery across injected shift cardinalities and random seeds.}
\textbf{(A)} Inclusion frequency of each feature in the identified shift set, aggregated across random seeds, as a function of the localized domain-shift cardinality.
\textbf{(B)} Pruned-to-injected domain-shift ratio as a function of the cardinality of the injected localized shift. The solid line denotes the mean across seeds, the shaded region denotes the 25th--75th percentile range across seeds, and the dashed horizontal line marks the ideal recovery level.
\textbf{(C)} MLP two-sample classifier baseline in the full 20-dimensional space.
Target samples are ranked by the MLP density-ratio score, and the top-400 ranked samples are used as candidate anchors.
\(\mathrm{InjectedRecall@400}\) reports the fraction of injected localized-shift samples contained among these top-400 candidate anchors.
The solid line and shaded region denote the mean and 25th--75th percentile range across seeds, respectively.
}
    \label{fig:shift_set_robustness}
\end{figure*}

We next consider a localized discrepancy supported on a small subset of the data, a condition which would be challenging for other approaches. 
As shown in Fig.~\ref{fig:shift_set_robustness}, recovery improves sharply as the cardinality of the localized domain shift increases. For small injected populations ($100$--$150$ points), both the recovered feature sets and the pruned-to-injected ratio remain variable across random seeds, indicating that the signal is still weak relative to the background. From roughly $250$ points onward, however, the method enters a stable regime: panel A shows that the true supporting coordinates $(2,4,6,8,9)$ dominate the identified shift set with high inclusion frequency, while spurious coordinates are progressively eliminated. Consistently, panel B shows that the mean pruned-to-injected ratio rises from a low-recovery regime at small cardinalities to about $0.9$ at $250$ points and to near-ideal values above $0.95$ for cardinalities of $300$ and higher. Inter-seed variability follows the same trend, being largest at low cardinalities and markedly reduced once the localized shift is sufficiently represented.

As a representative classifier-based baseline, we also train an MLP two-sample discriminator following the classifier two-sample testing paradigm~\cite{lopez2017revisiting}.
Target samples are ranked by the resulting MLP density-ratio score, and the top-ranked samples are evaluated as candidate anchors.
Panel C reports \(\mathrm{InjectedRecall@400}\), the fraction of injected localized-shift samples contained among the top-400 MLP-ranked target samples.
The low recall indicates that the classifier-based ranking does not recover a sufficiently representative anchor set for this localized discrepancy.
Together, these results show that, once the shifted population exceeds an initial critical mass, the workflow can stably recover both the feature support of a localized domain shift and a pruned representative set of the samples associated with it.
In contrast, a standard classifier-based approach does not recover a sufficiently representative candidate-anchor set.

\subsection{ECG dataset: PTB-XL+ with 12SL features}
\label{sec:ecg_dataset}

We next evaluate the proposed workflow in a real clinical feature space, focusing on domain shifts among nominally healthy ECG cohorts. We use PTB-XL+, an extension of the publicly available PTB-XL dataset of 12-lead ECG recordings~\cite{PTBxl,PTBxlplus}. PTB-XL+ combines ECG-derived feature representations, diagnostic statements, and acquisition metadata, making it well suited for constructing clinically controlled cohorts while auditing non-clinical sources of variation.
In this work, we focus on the Marquette 12SL feature set. Each recording is represented by a 782-dimensional vector of clinically interpretable ECG features, including global and lead-specific amplitudes, intervals, and morphology-related measurements. Prior to all analyses, features are standardized to zero mean and unit variance so that differences in physical units and dynamic ranges do not dominate distance computations.
We use diagnostic statements and metadata to restrict the analysis to nominally healthy recordings and to construct cohort pairs matched for age and sex composition while varying measurement-device composition. Only the standardized 12SL feature vectors are provided to the method; diagnostic statements and metadata are used exclusively for cohort construction and post hoc interpretation.
We use this design to test whether the proposed workflow can detect device-induced domain shifts in healthy ECG cohorts, localize the samples carrying these shifts, and identify active feature subspaces associated with the detected discrepancies. This experiment therefore complements the synthetic benchmark by replacing known feature-level ground truth with a controlled real-world source of variation. 
%
%
\begin{figure*}[t]
    \centering
    \includegraphics[width=.9\textwidth]{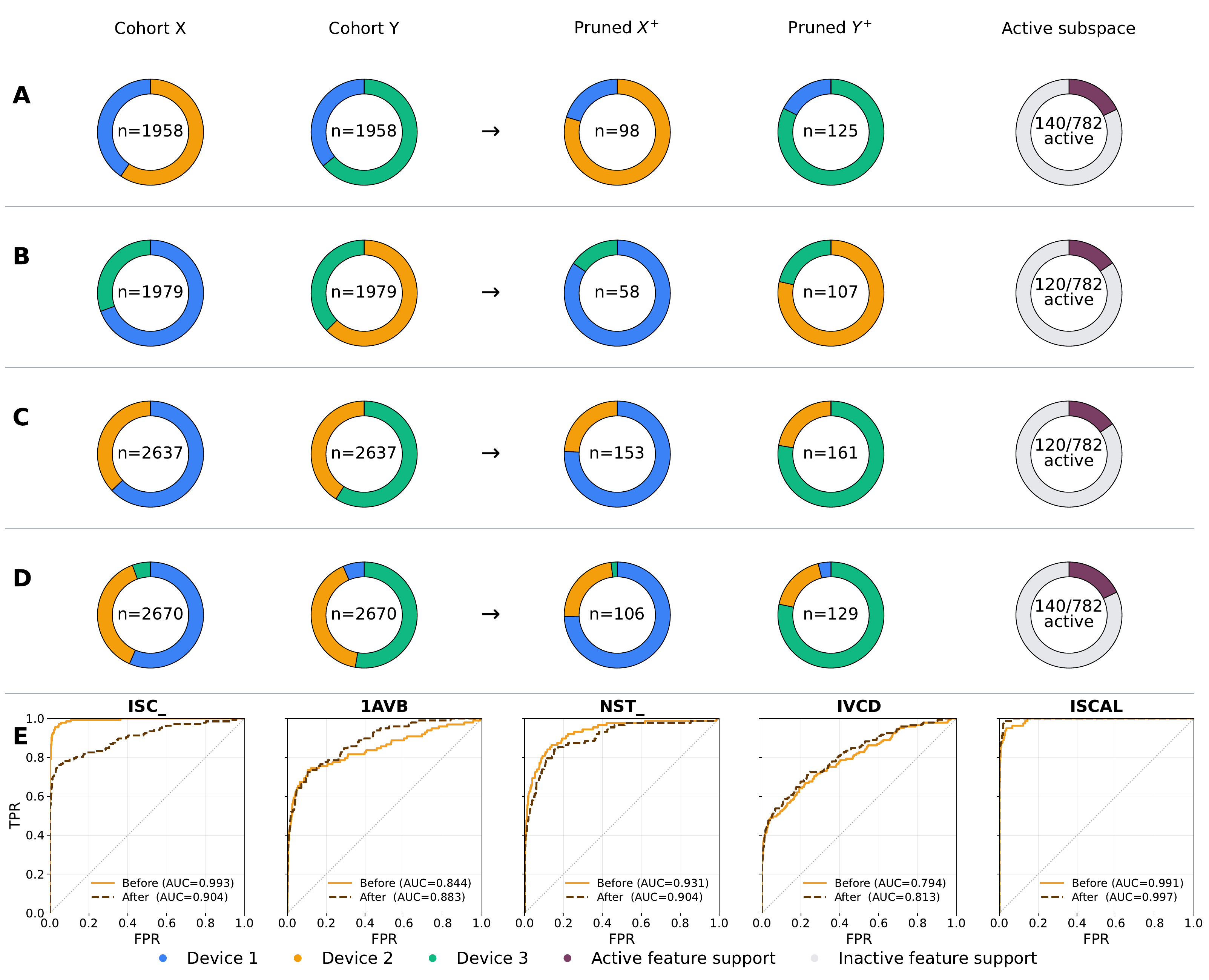}
    \caption{
    \textbf{Device-induced domain shifts in healthy ECG cohorts.}
    Each row corresponds to one controlled comparison between two age- and
    sex-matched healthy ECG cohorts, $\mathcal{X}$ and $\mathcal{Y}$,
    constructed from three measurement-device-specific subsets.
    Cases A--C are primary device-composition contrasts, each with one device
    specific to $\mathcal{X}$, one device approximately shared between
    cohorts, and one device specific to $\mathcal{Y}$.
    Case D is a perturbed version of Case~C, preserving the same dominant
    contrast while introducing a small fraction of the opposite device into
    each cohort.
    The first two columns show the device composition of the input cohorts.
    The next two columns show the device composition of the pruned
    representative sets $\hat{\mathcal{X}}^{+}$ and
    $\hat{\mathcal{Y}}^{+}$ extracted by our procedure.
    The active-subspace column reports the size of the feature support
    selected by Neighbor-Enriched subspace localization using
    $\hat{\mathcal{Y}}^{+}$ as query samples.
    Row~E shows AUCROC curves for binary NORM-vs-PATH classification on five
    illustrative pathologies, evaluated before and after equalization
    under the Case~C device-composition split.
    The classifier shown in Row~E was selected as the one with the largest
    mean absolute relative AUCROC change across the five tasks; all classifiers
    and the corresponding confusion matrices are reported in
    Supp.~\ref{supp:ECG_classification} and
    Figures~\ref{fig:ecg_classification1}--\ref{fig:ecg_classification2}.
    }
    \label{fig:ecg_device_shift}
\end{figure*}

 We analyzed  recordings on patients which are all classified as healthy acquired with three measurement devices: \texttt{AT-6 C 5.5}, \texttt{CS-12 E}, and \texttt{CS-12}. For readability, we refer to these devices as \texttt{Device 1}, \texttt{Device 2}, and \texttt{Device 3}, respectively. We then constructed controlled pairs of cohorts differing in their measurement-device composition. For each comparison, two cohorts $\mathcal{X}$ and $\mathcal{Y}$ were sampled from these device-specific subsets according to prescribed mixture proportions and then matched for age and sex. Device identity was used only to construct the cohorts and to interpret  the results post hoc.
We considered four device-composition comparisons (see Figure \ref{fig:ecg_device_shift}). Cases A--C correspond to distinct primary device-mixture contrasts. In each of these designs, one device is fully assigned to one cohort, a second device is split approximately equally between the two cohorts, and the third device is absent from that cohort but fully assigned to the opposite cohort. Thus, each comparison contains one shared device component and two device-specific components, creating a controlled device-composition contrast while preserving a partially common acquisition background. Case D is a perturbed version of case C, in which the same dominant contrast is preserved but each cohort contains a small fraction of the device that is dominant in the opposite cohort. This design allows us to ask whether the method responds coherently to acquisition-related distributional changes: if the detected discrepancy is device-related, the pruned representative samples should be enriched in the device components that are imbalanced between the two cohorts.

Figure~\ref{fig:ecg_device_shift} summarizes the results. Across comparisons, the pruned representative sets $\hat{\mathcal{X}}^{+}$ and $\hat{\mathcal{Y}}^{+}$ contain roughly $3\%$--$6\%$ of the corresponding healthy cohort, indicating that the detected domain shifts involve a non-negligible subset of recordings in each direction. The pruned sets  are systematically enriched in the device components that define the corresponding cohort mismatch. Thus, the procedure does not simply report a generic difference between healthy cohorts, but extracts representative samples associated with the device-induced domain shift.
The last column of Fig.~\ref{fig:ecg_device_shift} reports the size of the active feature support selected in this $\hat{\mathcal{Y}}^{+}$-query subspace. The selected supports are relatively broad, ranging from $120/782$ to $140/782$ features across comparisons. Within this attribution, the discrepancy is therefore represented by a distributed active support rather than by a handful of isolated ECG descriptors. This is an important advantage of our approach: it allows identifying domain shifts when they are supported in very high-dimensional subspaces. 
To quantify the stability of the selected feature supports, we computed pairwise Jaccard overlaps between the active supports obtained from the $\hat{\mathcal{Y}}^{+}$-query subspaces across the four comparisons. The perturbed case D showed strong overlap with case C ($J=0.86$), consistent with its construction as a moderate perturbation of the same dominant device contrast. Cases A, C, and D also showed substantial overlap ($J=0.79$--$0.80$), whereas case B had markedly lower overlap with the other comparisons ($J=0.24$--$0.26$). Taken together, these results show that controlled changes in measurement-device composition induce detectable, directionally localized domain shifts in healthy ECG cohorts, and that the corresponding feature supports reflect the similarity of the underlying device-composition contrasts.

To assess whether device-associated equalization can affect downstream
prediction, we performed a diagnostic classification check under the
Case~C device-composition split.
Five PTB-XL+ pathology labels were used as illustrative NORM-vs-PATH tasks:
1AVB, NST\_, IVCD, ISC\_, and ISCAL.
EagleEye equalization was estimated on the healthy cohorts only;
pathology-positive recordings were then added a posteriori according to the
same device-composition design, with training samples matched to
$\mathcal{X}$ and test samples matched to $\mathcal{Y}$.
Three classifiers spanning linear and nonlinear models were evaluated before
and after equalization (LR-L2, LR-EN, HGBT; see
Supp.~\ref{supp:ECG_classification} for details); row~E of Figure~3 reports
ROC curves for the classifier with the largest mean absolute relative AUCROC
change across the five tasks, with all conditions provided in
Figures~\ref{fig:ecg_classification1}--\ref{fig:ecg_classification2}.
The variation in the reference set of healthy subjects bring in some cases to better predictions (1AVB),  in other cases  to worse predictions (ISC), but in all the cases to \emph{different} predictions. This indicates, that even a highly curated dataset such as the one considered in this example small  domain shifts are present, and those domain shifts affect the classification results. 

\section{Discussion and limitations}
\label{sec:discussion}
The experiments presented above show that the proposed framework can detect distributional discrepancies between unlabeled datasets, attribute them to interpretable feature subspaces, and construct pruned/equalized subsets, both in controlled benchmarks with known ground truth and in a real clinical feature space. Several limitations should be considered when interpreting these results.

The equalized subsets $\mathcal{X}_\mathrm{eq}$ and $\mathcal{Y}_\mathrm{eq}$ produced by our procedure satisfy a statistical stopping criterion: the one-sided KS test on the anomaly-score tail no longer rejects at the prescribed significance level $\alpha$. This means that the detected excess mass has been reduced under the local scoring backbone, but it does not imply that the two equalized subsets are drawn from identical distributions. Residual discrepancies below the sensitivity threshold may remain. Equalization should therefore be interpreted as reduction of the detected distributional discrepancy, not as a certificate of distributional equivalence.
A related limitation is that the pipeline is triggered by the  local anomaly score estimated in the initial feature space. If no discrepancy is detected in the full feature space, the  Neighbor-Enriched Subspace Localization has no anchor set from which to start. This can occur in very high-dimensional settings when a weak discrepancy is confined to a small feature subset. In such cases,  our approach could be used in a hypothesis-driven mode: given a candidate set of data labelled as anomalous, one can attempt to identify the  subspace which would make the  anomaly statistically relevant (and reject the hypothesis if the subspace cannot be identified). This will be the object of future research. Also the derivation of formal convergence guarantees remain important directions for future work.

The MLP two-sample discriminator used in Section~\ref{sec:20Dbenchmark} is a representative classifier-based baseline following the C2ST paradigm~\citep{lopez2017revisiting}; through its posterior-odds score, it also provides a classifier-based density-ratio ranking that can be directly evaluated as a candidate-anchor selector. This comparison is not exhaustive. Future work could compare against other pointwise scoring baselines, including direct density-ratio estimators such as KLIEP~\citep{sugiyama2008direct}, after adapting them to the same anchor-recovery evaluation.

The ECG cohorts in Section~\ref{sec:ecg_dataset} are matched for age and sex, but residual confounding may remain from variables not recorded or not controlled in PTB-XL+, such body habitus, electrode placement protocol, or other acquisition-site effects. The detected discrepancies are consistent with device-associated acquisition effects, and the coherence of the feature supports across controlled device-composition contrasts supports this interpretation. However, the results should be read as identifying distributional discrepancies associated with measurement-device composition, not as proving device effect as their unique cause.

\section*{Code Availability}

The  code used to generate the results of this study will be made available upon acceptance for publication at the following GitHub repository: \url{https://github.com/sspring137/EagleEye}.

\newpage

{\small
\bibliographystyle{unsrtnat}
\bibliography{references}
}

\newpage
\appendix
\section{Supplementary Information}
\label{sec:supplement}

\subsection{Supplement:Tail-Based Bidirectional Equalization}
\label{sec:Supplement_IDE-ZooM}

\begin{algorithm}[h]
\caption{Tail-Based Bidirectional Equalization }
\label{alg:idezoom}
\small
\begin{algorithmic}[1]
\REQUIRE Samples $\mathcal{X},\mathcal{Y}$, neighborhood depth $K_M$, tail quantile level $q_{\mathrm{tail}}$, KS significance level $\alpha$
\STATE Initialize equalized samples $\mathcal{X}_{\mathrm{eq}}\leftarrow \mathcal{X}$, $\mathcal{Y}_{\mathrm{eq}}\leftarrow \mathcal{Y}$
\STATE Initialize pruned representative sets $\hat{\mathcal{X}}^{+}\leftarrow \emptyset$, $\hat{\mathcal{Y}}^{+}\leftarrow \emptyset$
\STATE Estimate the two backbone null score distributions for $\mathcal{Y}$- and $\mathcal{X}$-overdensities
\REPEAT
    \STATE Compute anomaly scores on all points in $\mathcal{X}_{\mathrm{eq}}$ and $\mathcal{Y}_{\mathrm{eq}}$ as in Section~\ref{sec:eagleeye_backbone}
    \STATE Set the tail thresholds $\tau_{\mathrm{tail}}^{(\mathcal{Y}^+)}$ and $\tau_{\mathrm{tail}}^{(\mathcal{X}^+)}$ from the corresponding null quantiles at level $q_{\mathrm{tail}}$
    \STATE Define the current tail candidates
    \[
    \mathcal{C}^{(\mathcal{Y}^+)} \leftarrow \{\mathcal{Y}_i\in\mathcal{Y}_{\mathrm{eq}}:\Upsilon_i^{(\mathcal{Y}^+)}\ge \tau_{\mathrm{tail}}^{(\mathcal{Y}^+)}\},
     \ \ \ \  
    \mathcal{C}^{(\mathcal{X}^+)} \leftarrow \{\mathcal{X}_j\in\mathcal{X}_{\mathrm{eq}}:\Upsilon_j^{(\mathcal{X}^+)}\ge \tau_{\mathrm{tail}}^{(\mathcal{X}^+)}\}
    \]
    \STATE Run one-sided two-sample KS tests comparing the empirical tails of $\mathcal{C}^{(\mathcal{Y}^+)}$ and $\mathcal{C}^{(\mathcal{X}^+)}$ against the corresponding null tails
    \WHILE{at least one of the two KS tests rejects at level $\alpha$}
        \IF{$\mathcal{C}^{(\mathcal{Y}^+)}$ is active}
            \STATE Select the most extreme point(s) in $\mathcal{C}^{(\mathcal{Y}^+)}$
            \STATE Let $\mathcal{N}^{(\mathcal{Y}^+)} \subset \mathcal{Y}_{\mathrm{eq}}$ denote the selected point(s) together with all neighbors up to the first point belonging to $\mathcal{X}_{\mathrm{eq}}$
            \STATE Update $\hat{\mathcal{Y}}^{+}\leftarrow \hat{\mathcal{Y}}^{+}\cup \mathcal{N}^{(\mathcal{Y}^+)}$ and $\mathcal{Y}_{\mathrm{eq}}\leftarrow \mathcal{Y}_{\mathrm{eq}}\setminus \mathcal{N}^{(\mathcal{Y}^+)}$
        \ENDIF
        \IF{$\mathcal{C}^{(\mathcal{X}^+)}$ is active}
            \STATE Select the most extreme point(s) in $\mathcal{C}^{(\mathcal{X}^+)}$
            \STATE Let $\mathcal{N}^{(\mathcal{X}^+)} \subset \mathcal{X}_{\mathrm{eq}}$ denote the selected point(s) together with all neighbors up to the first point belonging to $\mathcal{Y}_{\mathrm{eq}}$
            \STATE Update $\hat{\mathcal{X}}^{+}\leftarrow \hat{\mathcal{X}}^{+}\cup \mathcal{N}^{(\mathcal{X}^+)}$ and $\mathcal{X}_{\mathrm{eq}}\leftarrow \mathcal{X}_{\mathrm{eq}}\setminus \mathcal{N}^{(\mathcal{X}^+)}$
        \ENDIF
        \STATE Update $\mathcal{C}^{(\mathcal{Y}^+)} \leftarrow \mathcal{C}^{(\mathcal{Y}^+)} \cap \mathcal{Y}_{\mathrm{eq}}$ and $\mathcal{C}^{(\mathcal{X}^+)} \leftarrow \mathcal{C}^{(\mathcal{X}^+)} \cap \mathcal{X}_{\mathrm{eq}}$
        \STATE Recompute anomaly scores only for points currently belonging to $\mathcal{C}^{(\mathcal{Y}^+)}$ and $\mathcal{C}^{(\mathcal{X}^+)}$, then re-evaluate the two KS tests on the updated tails
    \ENDWHILE
\UNTIL{both KS tests do not reject after global recomputation}
\RETURN $\hat{\mathcal{X}}^{+},\hat{\mathcal{Y}}^{+},\mathcal{X}_{\mathrm{eq}},\mathcal{Y}_{\mathrm{eq}}$
\end{algorithmic}
\end{algorithm}

\subsection{ Supplement: Neighbor-Enriched feature selection (implementation details)}
\label{sec:supp_c1}

\paragraph{S1: Inputs, target/query definition, and defaults.}
C1 operates on a feature matrix $\mathcal{U}\in\mathbb{R}^{n\times d}$ (rows are samples, columns are features),
which in our experiments is obtained by pooling the two cohorts used by EagleEye. We assume $\mathcal{U}$ is standardised per feature (zero mean and unit variance) before running feature selection (\emph{in our experiments we perform this preprocessing step once and reuse it throughout}).

The neighbour-enrichment objective is defined through two boolean masks of length $n$:
\begin{itemize}
\item \texttt{target\_mask}$\in\{0,1\}^n$: \texttt{target\_mask}[j]=1 if sample $j$ is a desired neighbour (a ``hit'').
\item \texttt{query\_mask}$\in\{0,1\}^n$: \texttt{query\_mask}[i]=1 if sample $i$ is a query/anchor whose
      neighbourhood we aim to enrich in target samples.
\end{itemize}

In the $\mathcal{X}/\mathcal{Y}$ setting used throughout the paper, we typically set the target set to the test
cohort ($\mathcal{T}=\mathcal{Y}$), i.e.\ \texttt{target\_mask} marks the rows of the pooled matrix that belong to $\mathcal{Y}$, and set the query set $\mathcal{Q}\subseteq\mathcal{Y}$ to the pruned returned by EagleEye (so \texttt{query\_mask} marks only those selected $\mathcal{Y}$ rows). If \texttt{query\_mask}
is not provided, the implementation defaults to \texttt{query\_mask}=\texttt{target\_mask}. 

Unless stated otherwise we use the following defaults:
$K=100$ neighbours, $\beta=0.2$ (rank tie-break strength), $T=3000$ epochs, mini-batch size $B=200$, stratified query fraction \texttt{pos\_frac}$=0.5$, temperature annealing $\tau_{\text{start}}=1.0\rightarrow\tau_{\text{end}}=0.1$ (geometric),
Adam learning rate \texttt{lr}$=10^{-2}$, $\ell_1$ penalty \texttt{l1}$=10^{-2}$,
$5$-fold cross-validation (\texttt{n\_splits}$=5$), and random seed $0$. 

Mini-batches are stratified by \texttt{query\_mask}: \texttt{pos\_frac} is the fraction of \emph{query} samples
included in each batch (default $0.5$).


\paragraph{S2: Nonnegative weights and scale-normalised effective weights.}
We learn one nonnegative weight per feature via an unconstrained parameter vector
$\ell\in\mathbb{R}^d$ (stored as \texttt{logw} in the implementation), initialised to zeros.
The optimiser updates $\ell$; the weights $w_{\mathrm{raw}}$ and $w_{\mathrm{eff}}$ are deterministic functions of $\ell$.
At each optimisation step we map $\ell$ to \emph{raw} nonnegative weights through the Softplus transform,
\[
w_{\mathrm{raw},j}=\mathrm{softplus}(\ell_j)=\log\!\bigl(1+e^{\ell_j}\bigr)\quad\Rightarrow\quad
w_{\mathrm{raw}}\in\mathbb{R}^d_{\ge 0}.
\]
This guarantees nonnegativity at every iteration while remaining smooth for gradient-based optimisation; weights
can become arbitrarily small (effectively zero in practice) as $\ell_j\to -\infty$.

Because multiplying all weights by a common constant does not change neighbour identities up to a global rescaling
of distances, unconstrained optimisation can drift toward trivially large (or small) weights. We therefore define
\emph{effective} weights used in the distance metric by fixing the global scale to have sum approximately $d$:
\[
w_{\mathrm{eff}} \;=\; w_{\mathrm{raw}}\cdot \frac{d}{\sum_{j=1}^d w_{\mathrm{raw},j}}.
\]
Importantly, the implementation \emph{stops gradients} through the denominator (i.e., it treats the normalising sum
as a constant when computing parameter updates) and adds a small stabiliser:
\[
w_{\mathrm{eff}} \;=\; w_{\mathrm{raw}}\cdot \frac{d}{\mathrm{stopgrad}\!\left(\sum_{j=1}^d w_{\mathrm{raw},j}\right)+10^{-8}}.
\]

The resulting weights satisfy $\sum_j w_{\mathrm{eff},j}\approx d$, up to the $10^{-8}$ stabiliser.

Stopping gradients through the normalisation prevents the optimiser from compensating the sparsity penalty
(by shrinking the denominator) instead of genuinely driving many feature weights toward zero; it also makes the
effective weights comparable across runs because their overall ``budget'' is fixed. Distances and neighbour
probabilities are computed using $w_{\mathrm{eff}}$, while sparsity is enforced via an $\ell_1$ penalty on
$w_{\mathrm{raw}}$ (see S4).


\paragraph{S3: In-batch soft $K$NN neighbourhood probabilities.}
Given a mini-batch feature matrix $Z_b\in\mathbb{R}^{B\times d}$ (rows are samples), we compute neighbour
probabilities $P\in[0,1]^{B\times B}$ using the current effective weights $w_{\mathrm{eff}}$ (from S2).

We first rescale features elementwise and compute all pairwise squared Euclidean distances within the batch:
\[
\tilde Z_b = Z_b \odot w_{\mathrm{eff}},\qquad
d^2_{ij} = \|\tilde z_i-\tilde z_j\|_2^2,\qquad i,j\in\{1,\dots,B\}.
\]

We convert distances into similarity logits (unnormalised similarity scores) via
\[
\ell_{ij} = -\frac{d^2_{ij}}{\max(10^{-6},\tau)},
\]
so larger logits mean ``more similar''. The $\max(10^{-6},\tau)$ guard prevents numerical issues when $\tau$ is
very small, and self-matches are excluded by subtracting a large constant from the diagonal:
\[
\ell_{ii}\leftarrow \ell_{ii}-10^9.
\]

Neighbour probabilities are obtained by applying a row-wise softmax restricted to the $K$ largest logits in each
row. Concretely, for each $i$ we keep only the indices of the top-$K$ logits $\{\ell_{ij}\}_j$, set the remaining
entries to $-\infty$, and apply a softmax:
\[
P_{ij}=\frac{\exp(\ell_{ij})}{\sum_{j'\in \mathrm{TopK}(i)}\exp(\ell_{ij'})}\;\;\text{for}\;\; j\in\mathrm{TopK}(i),
\qquad
P_{ij}=0\;\;\text{otherwise}.
\]
If $K\ge B$, we default to a full row-wise softmax over all non-diagonal entries.

The resulting matrix $P$ satisfies $\sum_j P_{ij}=1$ for each row $i$ and can be interpreted as a soft assignment
distribution over the $K$ nearest neighbours of sample $i$ within the current mini-batch.
\paragraph{S4: Training objective and optimisation procedure.}
We learn feature weights by maximising \emph{soft target enrichment} of the query samples' neighbourhoods, using
mini-batch stochastic optimisation with Adam.

Let \texttt{query\_mask} mark the query set $\mathcal{Q}$ and let the remaining samples be \emph{non-queries}.
Each mini-batch corresponds to an index set $\mathcal{B}\subseteq \mathcal{U}$ of size $B$ and is constructed by
sampling, without replacement, approximately $( B\cdot\texttt{pos\_frac})$ query indices and
$(B- B\cdot\texttt{pos\_frac})$ non-query indices, then shuffling the concatenated indices. Across epochs we anneal the softmax temperature $\tau$ \emph{geometrically} from $\tau_{\text{start}}$ to
$\tau_{\text{end}}$.

Within a batch, we compute neighbour probabilities $P$ via the soft $K$NN construction in S3 and define, for each
batch sample $i$, the \emph{soft target mass}
\[
m_i \;=\; \sum_{j=1}^{B} P_{ij}\,\mathds{1}\{j \in \mathcal{T}\cap \mathcal{B}\},
\]
where the target indicator corresponds to \texttt{target\_mask} restricted to the current batch.
Only batch samples that are queries contribute to the objective. The batch objective is
\[
\mathrm{obj} \;=  \frac{1}{|\mathcal{Q}\cap \mathcal{B}|}\sum_{i\in\mathcal{Q}\cap \mathcal{B}} m_i\ ,
\]
(with the update skipped if $\mathcal{Q}\cap\mathcal{B}=\emptyset$).
We maximise the objective while promoting sparsity via an $\ell_1$ penalty on raw weights:
\[
\mathcal{L} \;=\; -\,\mathrm{obj} \;+\; \lambda\sum_{j=1}^{d} w_{\mathrm{raw},j},
\]
where $\lambda=10^{-2}$  and $w_{\mathrm{raw}}$ is the Softplus-transformed
nonnegative weight vector from S2.
We use Adam on the underlying unconstrained parameters with learning rate \texttt{lr}$=10^{-2}$ and default Adam hyperparameters. 


\paragraph{S5: From weights to a discrete feature subset.}
After training, we use the effective weights $w_{\mathrm{eff}}$ to rank features and to select the subset size
$m$ via a leakage-safe hard $K$NN cross-validation procedure (with the same $K$ used throughout the paper).

For each candidate subset size $m$, we take the indices of the $m$ largest weights,
$\mathrm{TopM}(w_{\mathrm{eff}},m)$. To speed up the search, the candidate values of $m$ are restricted to a
predefined grid. In the current implementation, the grid starts from $d$ and then sweeps down to $150$ in steps
of $50$, then continues down to $50$ in steps of $5$, then down to $16$ in steps of $2$, and finally considers
each integer from $15$ down to $1$.

To choose $m$, we evaluate each candidate subset $\mathrm{TopM}(w_{\mathrm{eff}},m)$ using $K$-fold cross-validation.
Here, ``training'' refers to the portion of the pooled dataset used to \emph{define neighbourhoods} for that fold,
while ``validation'' refers to held-out query samples used only to \emph{test} how target-enriched their
neighbourhoods are under the chosen subset.

Concretely, in each fold we split the pooled indices into a training set and a validation set. We then compute
hard $K$ nearest neighbours for each validation query sample using distances in the selected feature subset, but
we restrict the neighbour search to the training set only (validation samples are never used as neighbours of one
another). This avoids optimistic bias when assessing enrichment.

We split the data into $K$ folds in a stratified manner that preserves the query/non-query proportion across
folds and ensures that each fold contains both groups whenever possible.

For a validation query sample $i$, let $c_i$ be the number of target samples among its $K$ nearest neighbours, and
let $\phi_i=c_i/K$ denote the corresponding neighbourhood purity.

When several subsets yield similar purity, we prefer the one where the target neighbours tend to appear
\emph{earlier} (i.e., at smaller ranks, hence closer). To quantify this, we use a standard rank-discounted score
from information retrieval (discounted cumulative gain, DCG, and its normalized variant nDCG; 
\cite{JarvelinKekalainen2002_DCG,ManningRaghavanSchutze2008_IR}): each target neighbour contributes to the score,
but contributions are down-weighted as the neighbour rank increases. We normalize by the best possible value given
the same hit count $c_i$, yielding a normalized rank-sensitivity $q_i\in[0,1]$ with $q_i=1$ when the $c_i$ target
neighbours are the closest ones.

We then combine purity and rank sensitivity as
\[
s_i = \phi_i + \beta(1-\phi_i)q_i,
\]
and define the subset score as the mean of $s_i$ over validation queries, averaged across folds. In addition to
this main score (with the user-specified $\beta$), we also report a ``purity-only'' curve obtained by setting
$\beta=0$, which corresponds to the target fraction among hard $K$NN neighbours.

Finally, we select $m^\star$ as the value in the grid that maximizes the cross-validated mean score, and return
$\mathrm{TopM}(w_{\mathrm{eff}},m^\star)$ as the selected feature subset.

\subsection{Default hyperparameters}
\label{supp:default_hyperparameters}

Unless otherwise stated, all experiments use the default hyperparameters reported below. 
These values were kept fixed across the synthetic and ECG experiments, except when a parameter was explicitly varied as part of the experimental design.

\paragraph{EagleEye local two-sample scoring.}
\begin{itemize}
    \item Maximum neighbourhood depth: \(K_M = 400\) for the synthetic benchmark and \(K_M = 200\) for the ECG experiments.
    \item Null calibration: Monte Carlo calibration using Bernoulli sequences as in \citep{EagleEye26}.
\end{itemize}

\paragraph{Clustering}
\begin{itemize}
    \item We use Density Peaks Advanced clustering with default parameters as in  \citep{EagleEye26}.
\end{itemize}

\paragraph{Bidirectional tail-KS equalization.}
\begin{itemize}
    \item Tail threshold: \(q_{\mathrm{tail}} = 0.97\) null quantile.
    \item Tail test: one-sided two-sample Kolmogorov--Smirnov test comparing observed and null anomaly-score tails.
    \item KS significance level: \(\alpha = 0.05\), unless otherwise stated.
    \item Pruning rule: pruning is activated on a side only when the corresponding one-sided KS tail test rejects at level \(\alpha\).
    \item Pruned neighbourhood: for each selected tail point, samples are removed up to the first neighbour belonging to the opposite dataset.
    \item Recalculation strategy: anomaly scores are updated locally within the active tails during pruning and recomputed globally after both tail tests stop rejecting.
    \item Stopping rule: the procedure stops when both one-sided KS tail tests no longer reject after global recomputation.
\end{itemize}

\paragraph{Neighbor-Enriched Subspace Localization.}
\begin{itemize}
    \item Number of neighbours in the enrichment objective: \(K = 100\).
    \item Feature weights: one nonnegative raw weight per feature, transformed through a softplus parameterization.
    \item Effective weight normalization: effective weights are rescaled to have total weight approximately equal to the ambient dimension \(d\).
    \item Sparsity penalty: \(\lambda = 10^{-2}\) on raw feature weights.
    \item Optimizer: Adam.
    \item Learning rate: \(10^{-2}\).
    \item Number of epochs: \(3000\).
    \item Mini-batch size: \(200\).
    \item Query fraction in stratified mini-batches: \(0.5\).
    \item Temperature schedule: geometric annealing from \(\tau_{\mathrm{start}} = 1.0\) to \(\tau_{\mathrm{end}} = 0.1\).
    \item Model selection: leakage-safe hard-KNN cross-validation over candidate subset sizes.
    \item Number of cross-validation folds: \(5\).
    \item Rank-discount strength for subset scoring: \(\beta = 0.2\).
    \item Random seed: \(0\), unless otherwise stated.
\end{itemize}

\paragraph{MLP two-sample classifier baseline.}
\begin{itemize}
    \item Task: binary discrimination between samples from \(\mathcal{X}\) and \(\mathcal{Y}\).
    \item Input space: full 20-dimensional feature space for the synthetic localized-shift benchmark.
    \item Architecture: fully connected multilayer perceptron with hidden layers \((128,64,32)\).
    \item Activation: ReLU.
    \item Optimizer: Adam.
    \item Learning rate: \(10^{-3}\).
    \item \(L_2\) regularization parameter: \(10^{-4}\).
    \item Batch size: \(256\).
    \item Maximum number of training iterations: \(300\).
    
    \item Density-ratio score: target samples are ranked by the classifier posterior-odds score,
    \[
        \hat r(x) = \frac{\hat p(y=\mathcal{Y}\mid x)}{1-\hat p(y=\mathcal{Y}\mid x)}.
    \]
    \item Anchor-recovery evaluation: \(\mathrm{InjectedRecall@400}\), defined as the fraction of injected localized-shift samples contained among the top-400 ranked target samples.
\end{itemize}

\subsection{Controlled 20D benchmark construction}
\label{supp:20d_benchmark_construction}

We constructed a controlled 20-dimensional benchmark to evaluate recovery of both sample-level domain shifts and their supporting feature coordinates under known ground truth. The background distribution is a four-component Gaussian mixture in \(\mathbb{R}^{20}\). Unless otherwise stated, each reference cohort contains \(50{,}000\) background samples.

\paragraph{Background Gaussian mixture.}
The feature space is \(d=20\). The reference distribution is
\[
    P_0(x) = \sum_{m=1}^{4} \pi_m \mathcal{N}(x;\mu_m,\Sigma_m),
\]
with mixture weights
\[
    \pi = (0.35, 0.30, 0.20, 0.15).
\]
The component means are
\[
\begin{aligned}
\mu_1 &= (0.0,0.0,0.5,-0.5,0.0,0.3,0.0,0.0,0.2,0.0,0.0,0.0,0.0,0.0,0.0,0.0,0.0,0.0,0.0,0.0),\\
\mu_2 &= (2.5,-1.0,-0.5,1.0,0.5,-0.3,0.0,0.2,-0.2,0.0,0.0,0.1,0.0,0.0,0.0,0.0,0.0,0.0,0.0,0.0),\\
\mu_3 &= (-2.0,1.5,0.0,0.5,-1.0,0.0,0.3,-0.2,0.0,0.0,0.2,0.0,0.0,0.0,0.0,0.0,0.0,0.0,0.0,0.0),\\
\mu_4 &= (1.0,2.0,-1.0,-1.0,0.0,0.5,-0.5,0.0,0.0,0.3,0.0,0.0,0.0,0.0,0.0,0.0,0.0,0.0,0.0,0.0).
\end{aligned}
\]
All covariance matrices are diagonal. Their diagonal entries are
\[
\begin{aligned}
\mathrm{diag}(\Sigma_1) &= (1.2,1.0,0.8,0.9,0.7,0.8,1.0,1.0,0.9,1.0,1.0,1.0,1.0,0.8,0.8,0.8,0.9,0.9,0.9,0.9),\\
\mathrm{diag}(\Sigma_2) &= (0.9,1.1,0.7,0.8,0.8,0.7,1.0,0.9,1.0,1.0,1.0,0.9,1.0,0.8,0.8,0.8,1.0,1.0,0.9,0.9),\\
\mathrm{diag}(\Sigma_3) &= (1.0,0.8,1.0,0.9,0.7,1.1,0.8,0.9,1.0,1.0,0.9,1.0,1.0,0.9,0.8,0.8,0.8,0.9,0.9,1.0),\\
\mathrm{diag}(\Sigma_4) &= (1.1,0.9,0.8,1.0,0.8,0.8,0.9,1.0,1.0,0.9,1.0,1.0,0.9,0.8,0.8,0.8,0.9,0.9,1.0,1.0).
\end{aligned}
\]

\paragraph{Global domain-shift experiment.}
The global domain-shift experiment introduces a systematic perturbation in one background mixture component. Specifically, component \(m=2\) is shifted in the second cohort along coordinates
\[
    \{0,1,3\}.
\]
The displacement magnitude is controlled by the scalar parameter \(\delta\), denoted by \(\sigma\) in the main text. We vary
\[
    \sigma \in \{0,0.05,0.10,\ldots,0.50\}.
\]
For each value of \(\sigma\), the analysis is repeated across \(11\) random seeds. 

\paragraph{Localized domain-shift experiment.}
The localized domain-shift experiment injects a compact minority population into one cohort. The injected population is associated with component \(m=1\) and is supported on the ground-truth feature set
\[
   \{2,4,6,8,9\}.
\]
The number of injected samples is varied as
\[
\{100,150,200,250,300,350,400,450,500\}.
\]
Injected samples are generated around an anchor defined relative to the standard deviations of the selected background component. Along the active dimensions \(S_{\mathrm{local}}\), the anchor offset in units of component standard deviation is
\[
    (1.1,-0.4,0.4,-0.2,0.3).
\]
The injected cluster has anisotropic geometry with scales, again expressed in units of the component standard deviations,
\[
    (0.11,0.08,0.04,0.04,0.03),
\]
and a random rotation is applied within the active subspace. In ambient inactive dimensions, the injected samples follow the corresponding background component with shrinkage factor \(0.7\).


\subsection{ECG dataset: classification}
\label{supp:ECG_classification}

\begin{figure*}[t]
    \centering
    \includegraphics[width=.9\textwidth]{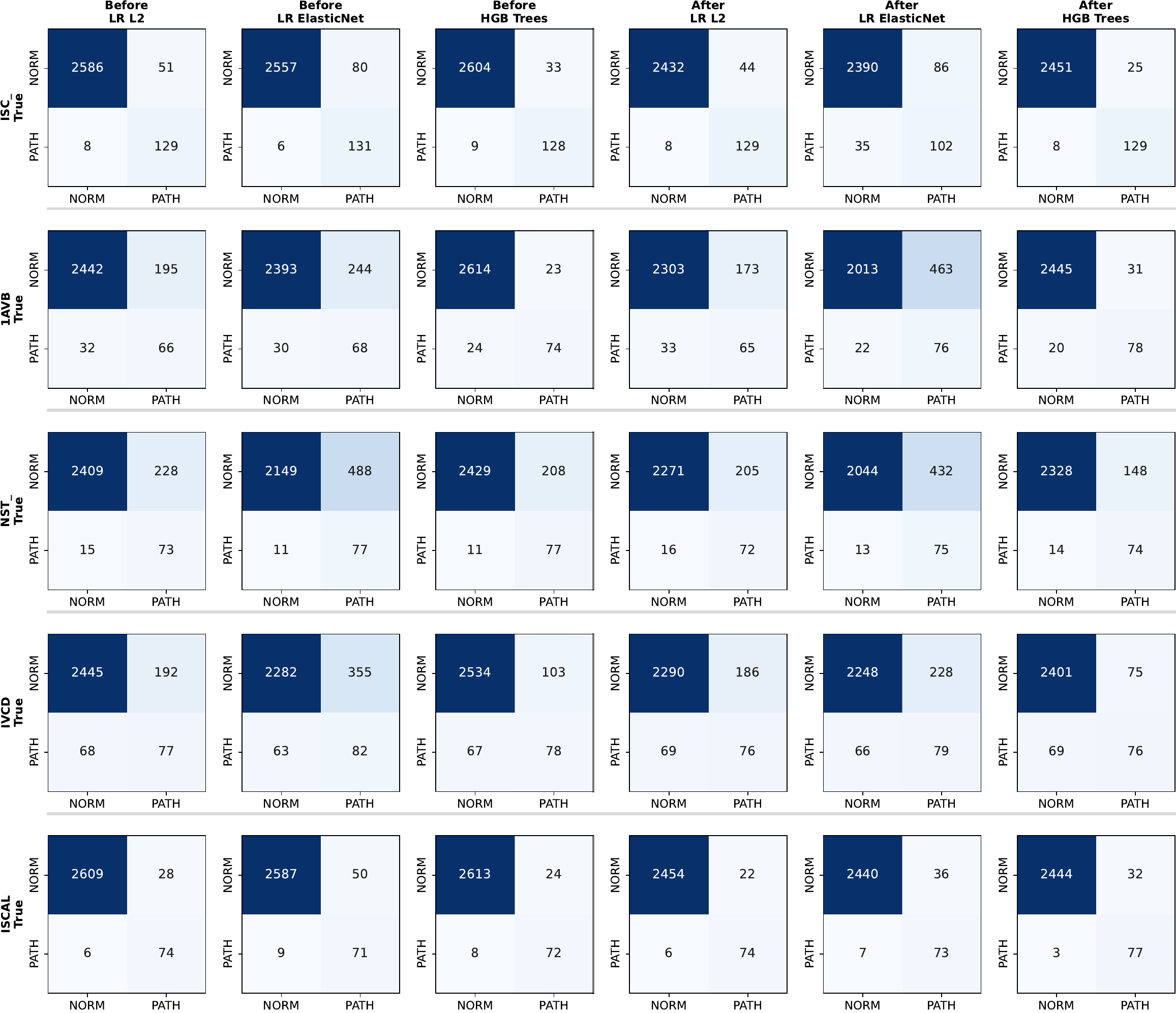}
    \caption{%
        \textbf{Confusion matrices for NORM-vs-PATH binary classification
        before and after EagleEye equalization.}
        Rows correspond to the five illustrative pathology tasks
        (ISC\_, 1AVB, NST\_, IVCD, ISCAL);
        columns correspond to six conditions:
        three classifiers (LR-L2, LR-EN, HGBT) evaluated before equalization
        (left block) and after equalization (right block).
        Each $2{\times}2$ matrix reports prediction counts with predicted class on the horizontal axis and true class on the vertical axis, using NORM and PATH labels. Cohorts are constructed under the
        Case~C device-composition split; see
        Section~\ref{supp:ECG_classification} for classifier details.
    }
    \label{fig:ecg_classification1}
\end{figure*}
To assess whether device-associated equalization can affect downstream prediction, we performed a diagnostic classification check under the same
Case~C device-composition split used in the ECG equalization experiment.
We considered five PTB-XL+ pathology labels, used here as illustrative NORM-vs-PATH classification tasks: first-degree atrioventricular block
(1AVB), nonspecific ST changes (NST\_), intraventricular conduction
disturbance (IVCD), ischemic ST-T changes (ISC\_), and ischemic anterolateral ST-T changes (ISCAL). The
782-dimensional Marquette 12SL feature vectors were used as input after the standardization described in Section~3.2 of the main text, with no additional
task-specific preprocessing.

We evaluated three classifiers. \textit{(i)}~\emph{LR-L2}: logistic
regression with $\ell_2$ regularization
(\texttt{penalty=``l2''}, $C=0.01$, \texttt{solver=``lbfgs''},
\texttt{max\_iter=2000}), used as a ridge-regularized linear baseline.
\textit{(ii)}~\emph{LR-EN}: logistic regression with ElasticNet
regularization
(\texttt{penalty=``elasticnet''}, $C=0.01$, $\ell_1$\texttt{\_ratio=0.5},
\texttt{solver=``saga''}, \texttt{max\_iter=3000}), providing a sparse
linear baseline through the mixed $\ell_1$/$\ell_2$ penalty.
\textit{(iii)}~\emph{HGBT}: histogram gradient-boosted trees
(\texttt{max\_depth=3}, \texttt{learning\_rate=0.03},
\texttt{min\_samples\_leaf=30}, \texttt{max\_iter=500}), with early stopping
on a 15\% validation fraction
(\texttt{n\_iter\_no\_change=20}). The shallow depth and minimum leaf-size constraint were used to limit overfitting in the imbalanced binary tasks.

\begin{figure*}[t]
    \centering
    \includegraphics[width=.9\textwidth]{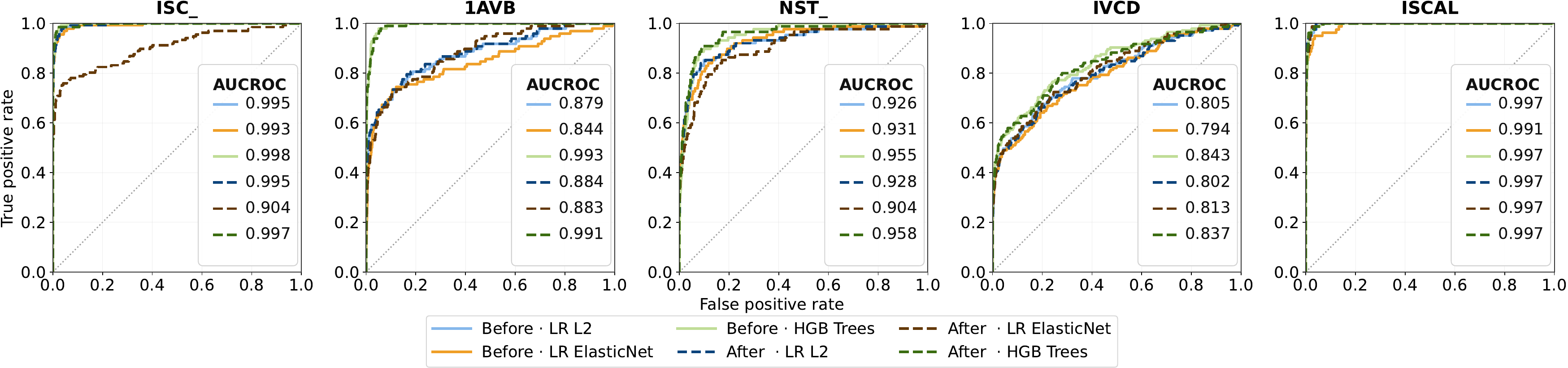}
    \caption{%
        \textbf{ROC curves for NORM-vs-PATH binary classification before
        and after EagleEye equalization.}
        Each panel corresponds to one of the five illustrative pathology
        tasks. Solid curves show performance before equalization; dashed
        curves show performance after equalization. Colors distinguish
        LR-L2, LR-EN, and HGBT, and AUCROC values are reported inside each
        panel. The classifier shown in row~E of Figure~3 of the main text
        is selected as the one with the largest mean absolute relative
        AUCROC change across the five pathology tasks; this supplementary
        figure reports all three classifiers for completeness. Cohorts are
        constructed under the Case~C device-composition split; see
        Section~\ref{supp:ECG_classification} for classifier details.
    }
    \label{fig:ecg_classification2}
\end{figure*}

Class imbalance was handled using inverse-frequency class weights computed on the training set,
$w_0 = 1$ and $w_1 = n_{\mathrm{NORM}} / n_{\mathrm{PATH}}$.
No oversampling or resampling was applied. EagleEye equalization was first estimated on the healthy cohorts of Case~C only. The downstream
NORM-vs-PATH tasks were then constructed a posteriori by adding pathology-positive recordings before and after equalization according to the
same device-composition design: pathology-positive training samples matched cohort $\mathcal{X}$, pathology-positive test samples matched cohort $\mathcal{Y}$, and samples
from the device shared by $\mathcal{X}$ and $\mathcal{Y}$ were split 50--50.  For each of the 5 cases, classifiers were trained on cohort $\mathcal{X}$ plus the corresponding
pathology-positive training samples and tested on cohort $\mathcal{Y}$ plus the
corresponding pathology-positive test samples, before and after equalization. 
Performance is summarized by ROC curves, AUCROC, and confusion matrices at the default 0.5 decision threshold. The full before/after results for all three classifiers are shown in
Figure~\ref{fig:ecg_classification2}, with the corresponding confusion
matrices reported in Figure~\ref{fig:ecg_classification1}.

\end{document}